\documentclass{article}

\PassOptionsToPackage{numbers, compress}{natbib}
\usepackage[preprint]{neurips_2024}

\usepackage[utf8]{inputenc} 
\usepackage[T1]{fontenc}    
\usepackage[backref=page]{hyperref}

\renewcommand*{\backref}[1]{}
\renewcommand*{\backrefalt}[4]{%
    \ifcase #1 %
    \or        (Cited on pg.~#2.)%
    \else      (Cited on pgs.~#2.)%
    \fi}
\usepackage{url}            
\usepackage{booktabs}       
\usepackage{amsfonts}       
\usepackage{nicefrac}       
\usepackage{microtype}      
\usepackage{xcolor}         
\usepackage{graphicx}
\usepackage{wrapfig}
\usepackage{lipsum}
\usepackage{amsmath}
\usepackage[capitalize,noabbrev,nameinlink]{cleveref}
\usepackage{amssymb}
\usepackage{algorithm}
\usepackage{algpseudocode}
\usepackage{setspace}
\usepackage{array}
\usepackage{enumitem}

\newcommand{\acro}[1]{\textsc{\MakeLowercase{#1}}}
\DeclareMathOperator*{\argmax}{arg\,max}

\Crefname{section}{Sect.}{Sects.}
\Crefname{equation}{Eq.}{Eqs.}
\Crefname{figure}{Fig.}{Figs.}
\Crefname{tabular}{Tab.}{Tabs.}
\Crefname{table}{Tab.}{Tabs.}
\Crefname{appendix}{Appx.}{Appxs.}
\Crefname{algorithm}{Alg.}{Algs.}
\Crefname{observation}{Obs.}{Obs.}

\newcommand{\result}[2]{\makebox[2.5em]{\hfill#1}\makebox[2.7em]{\hfill(#2)}}
\newcommand{\resultbold}[2]{\makebox[2.5em]{\hfill\textbf{#1}}\makebox[2.7em]{\hfill\textbf{(#2)}}}
\newcommand{\resultblueit}[2]{\makebox[2.5em]{\hfill\emph{\color{blue} #1}}\makebox[2.7em]{\hfill\emph{\color{blue} (#2)}}}

\pdfoutput=1

\title{Amortized nonmyopic active search \\ via deep imitation learning}

\author{%
Quan Nguyen
\hskip 30pt
Anindya Sarkar
\hskip 30pt
Roman Garnett \\
Washington University in St.\ Louis \\
\texttt{\{quan,anindya,garnett\}@wustl.edu}
}

\begin{document}

\maketitle

\begin{abstract}
Active search formalizes a specialized active learning setting where the goal is to collect members of a rare, valuable class.
The state-of-the-art algorithm approximates the optimal Bayesian policy in a budget-aware manner, and has been shown to achieve impressive empirical performance in previous work.
However, even this approximate policy has a superlinear computational complexity with respect to the size of the search problem, rendering its application impractical in large spaces or in real-time systems where decisions must be made quickly.
We study the amortization of this policy by training a neural network to learn to search.
To circumvent the difficulty of learning from scratch, we appeal to imitation learning techniques to mimic the behavior of the expert, expensive-to-compute policy.
Our policy network, trained on synthetic data, learns a beneficial search strategy that yields nonmyopic decisions carefully balancing exploration and exploitation.
Extensive experiments demonstrate our policy achieves competitive performance at real-world tasks that closely approximates the expert's at a fraction of the cost, while outperforming cheaper baselines.
\end{abstract}

\section{Introduction}

Many problems in science and engineering share a common theme where an agent searches for rare and valuable items in a massive database; examples include fraud detection, product recommendation, and drug and materials discovery.
The bottleneck of this procedure is often the cost of labeling, that is, determining whether a candidate is one of the targets of the search.
For instance, in product recommendation, labeling can involve presenting a customer with a product they might enjoy, at the risk of interrupting the customer's shopping experience and consequently losing sales; in drug discovery, it takes time-consuming computer simulations and/or expensive laboratory experiments to characterize a potential drug.
This labeling cost rules out exhaustive screening and motivates strategic exploration of the search space.
Active search (\acro{AS}) frames this problem in the language of adaptive experimental design, and aims to develop policies that iteratively choose data points to label to uncover as many valuable points as possible under a labeling budget.

\acro{AS} has been thoroughly studied in previous work and sophisticated search policies have been developed \cite{garnett2012bayesian,jiang2017efficient,jiang2018efficient,jiang2019cost}.
Of particular interest is the work of \citet{jiang2017efficient}, who derived the Bayesian optimal policy under the simplifying assumption that future experiments are chosen simultaneously in a batch.
Here, the number of future experiments is set to be the remaining labeling budget so that this remaining budget is actively accounted for during policy computation. 
The authors called the resulting policy \emph{efficient nonmyopic search} (\acro{ENS}), which can be viewed as a budget-aware approximation to the true optimal policy.
They showed this budget-awareness induces nonmyopic decisions that automatically balance between strategic exploration of the space and timely exploitation of regions that likely yield many targets, and ultimately achieve state-of-the-art (\acro{SOTA}) search performance across many tasks.
Although the aforementioned simplifying assumption, combined with aggressive pruning, allows \acro{ENS} to be feasibly applied to problems of considerable size (100\,000+ in \citet{jiang2017efficient,jiang2018efficient,jiang2019cost}, for example), the policy retains a \emph{superlinear} computational complexity.
This complexity poses a challenge in (1) deploying in real-time applications where decisions need to be made quickly and (2) scaling to large spaces.
For instance, a guided data discovery task \cite{monadjemi2022guided} in visual analytics combines a search algorithm with an interactive visualization to assist a user with their analytic goals in real time, and the time available for the search algorithm to run is thus severely constrained.
Modern recommender systems such as YouTube and Amazon must quickly search over millions of items to make recommendations for a large number of users \cite{dulac2015deep}; similarly, there exist databases with billions of synthesizable molecules acting as search spaces for drug discovery \cite{tingle2023zinc}.

We aim to alleviate the computational cost of budget-aware search by training a small, relatively shallow feedforward neural network to mimic the behavior of the \acro{SOTA}, expert policy \acro{ENS}; policy computation is thus amortized as we deploy the trained network as the search policy.
We train this policy using the imitation learning technique DA\acro{GGER} \cite{ross2011reduction}, which aids the goal of behavior cloning by iteratively querying the expert's actions on states encountered by the network being trained.
This procedure is done with small, synthetic search problems where \acro{ENS} is cheap to query.
We find that the trained policy network successfully learns a beneficial nonmyopic search behavior and, despite the synthetic training data, incurs only a minor decrease in search performance at real-world tasks, in exchange for much faster decision-making.
We showcase the usefulness of this computationally lightweight policy with a wide range of search problems spanning diverse applications, including drug discovery tasks of an unprecedented multi-million scale.

\section{Preliminaries}

We first introduce the problem setting, the active search framework, the \acro{SOTA} search policy and its computational complexity to motivate our goal of amortization.

\subsection{Active search and the optimal policy}
\label{sec:as_formulation}

An active search (\acro{AS}) problem is defined by a finite set of data points $\mathcal{X} \triangleq \{ x_i \}$, among which there exists a rare, valuable subset $\mathcal{T} \subset \mathcal{X}$.
We use the term ``targets'' to refer to the members of this valuable subset, which we wish to collect from the entire space $\mathcal{X}$.
We further use membership in $\mathcal{T}$ as the labels for the points in $\mathcal{X}$: $y_i \triangleq \mathbb{I} \, [ x_i \in \mathcal{T} ], \forall x_i \in \mathcal{X}$.
The targets are not known \emph{a priori}, but whether a specific data point $x_i$ is one can be determined by querying an oracle returning the requested label $y_i$.
This oracle models the process of performing laborious experiments to characterize the data point in question (e.g., suggesting a given product to a customer and observing their subsequent clicking behavior, or performing computer simulations and laboratory experiments on a candidate molecule for drug discovery).
We thus assume the oracle to be expensive to query.
Concretely, we assume we may only query the oracle $T$ times, where $T\ll n \triangleq | \mathcal{X} |$; $T$ can be viewed as our querying budget.
We denote the set of data points and their revealed labels as $\mathcal{D} = \{ (x_i, y_i) \}$, and the set of those queried up to iteration $t \leq T$ as $\mathcal{D}_t = \{ (x_i, y_i) \}_{i = 1}^t$.
The goal of \acro{AS} is to design a policy that sequentially selects which data points to query so as to find as many targets throughout the $T$ iterations of the search as possible.
To express this preference for maximizing the number of ``hits'' across different terminal data sets collected at the end of the search $\mathcal{D}_T$, we use the utility function $u(\mathcal{D}_T) = \sum_{y_i \in \mathcal{D}_T} y_i$, which simply counts the number of targets in $\mathcal{D}_T$.

Previous works on \acro{AS} have derived the optimal policy under Bayesian decision theory that finds the query with the highest expected terminal utility \cite{garnett2012bayesian,jiang2017efficient}.
First, we build a probabilistic classifier that outputs the probability an unlabeled point $x$ has a positive label given the observed data set $\mathcal{D}$, denoted as $\Pr (  y = 1 \mid x, \mathcal{D} )$.
Then, at iteration $t + 1$, having collected $\mathcal{D}_t$, we query the data point:
\begin{equation}
\label{eq:optimal}
x_{t + 1}^* = \argmax_{x_{i + 1} \in \mathcal{X} \setminus \mathcal{D}_t} \mathbb{E} \big[ u (\mathcal{D}_T) \mid x_{i + 1}, \mathcal{D}_t \big],
\end{equation}
where the expectation is with respect to the label $y_{i + 1}$, and the future queries $(x_{t + 2}, x_{t + 3}, \ldots, x_T)$ are similarly computed in this optimal manner.
While this computation can theoretically be achieved via dynamic programming \cite{bellman1957dynamic}, it has a complexity of $O \big( (2 \, n)^\ell \big)$, where $n$ is again the size of the search space and $\ell = T - t$ is the length of the decision-making horizon (i.e., the remaining querying budget).
This exponential blowup stems from the fact that Bayesian decision theory compels the optimal policy to reason about not only the possible labels of a putative query, but also how each possible label affects subsequent future queries.
This computation therefore cannot be realized in almost all practical scenarios except for the very last few iterations when $\ell$ is sufficiently small \cite{garnett2012bayesian}.
A common strategy that we can adopt here is to limit the depth of the lookahead in this computation, effectively pretending $\ell$ is indeed small.
The simplest version of this is obtained by setting $\ell = 1$: we assume we have only one query remaining, and the optimal decision becomes maximizing the expected marginal utility gain, which is equivalent to greedily querying the most likely target:
\begin{align}
\begin{split}
x_T^* & = \argmax_{x_T \in \mathcal{X} \setminus \mathcal{D}_{T - 1}} \mathbb{E} \Big[ u \big( \mathcal{D}_{T - 1} \cup \{ x_{T}, y_T \} \big) \mid x_T, \mathcal{D}_{T - 1} \Big] \\
& = \argmax_{x_T \in \mathcal{X} \setminus \mathcal{D}_{T - 1}} \Pr ( y_T = 1 \mid x_T, \mathcal{D}_{T - 1} ).
\end{split}
\end{align}
We call this greedy policy the \emph{one-step} policy, as it computes the one-step optimal decision.
Albeit extremely computationally efficient, one-step does not consider exploratory queries that may lead to future gains.
As a result, it tends to get stuck in small regions of targets, failing to discover larger target sets due to insufficient exploration.
Theoretically, \citet{garnett2012bayesian} even showed that by limiting its lookahead, a myopic policy could perform worse than a less myopic one \emph{by any arbitrary degree}.
Motivated by the potential of nonmyopia in \acro{AS}, \citet{jiang2017efficient} developed an approximation to the optimal policy that accounts for the remaining budget $\ell$, which we examine next.

\subsection{Nonmyopic search via budget-awareness}

To avoid the high cost of reasoning about the dependence among labels of future queries, \citet{jiang2017efficient} made the simplifying assumption that, after our next query, all remaining future queries are made at the same time in a batch.
Under this assumption, the future queries in the lookahead in \cref{eq:optimal}---to be optimally chosen to maximize expected terminal utility---can be quickly identified as the set of $(\ell - 1)$ most likely targets \cite{jiang2017efficient}.
Their policy \acro{ENS} thus estimates the value of each putative query $x_i$ with the expected utility of the union of $x_i$ and the top $(\ell - 1)$ unlabeled points that are adaptively selected based on each possible label $y_i$.
Again, as the number of future queries in this policy computation is set to exactly match the true length of the decision-making horizon, \acro{ENS} actively accounts for the remaining labeling budget when making its queries.
The authors demonstrated the benefits of this budget-awareness by showing that \acro{ENS} exhibits nonmyopic, exploratory behavior when the budget is large, and automatically transitions to more exploitative queries as search progresses.
This strategic exploration ultimately allows \acro{ENS} to outperform many search baselines including the one-step policy.

While the aforementioned batch assumption avoids an exponential blowup in computational complexity,
\acro{ENS} still incurs a considerable cost, especially under large values of $n$, the size of the search space.
A na\"ive implementation with a generic classifier has a complexity of $O \big( n^2 \log n \big)$.
The official implementation by \citet{jiang2017efficient}, on the other hand, uses a lightweight $k$-nearest neighbor (\acro{NN}) classifier that (reasonably) assumes a certain level of locality when probabilities $\Pr ( y \mid x, \mathcal{D} )$ are updated in light of new data.
This structure allows for a faster computation of the batch of future queries in \acro{ENS}'s lookahead, and brings the complexity down to $O \big( n \, (\log n + m \log m + T) \big)$, where $m$ is the largest degree of any node within the nearest neighbor graph corresponding to the $k$-\acro{NN} \cite{jiang2017efficient}.
Unfortunately, this reduced complexity still poses a substantial challenge in two scenarios commonly encountered in \acro{AS}: large search spaces (e.g., drug discovery) and settings where queries must be rapidly computed (recommender and other real-time systems).
We address this problem by training an estimator, specifically a neural network, to learn the mapping from possible candidate queries to the output of \acro{ENS}, thus amortizing policy computation; the next section details our approach.

\section{Amortizing budget-aware active search}
\label{sec:imitation_learning}

Our goal is to amortize search with a neural network, replacing the time-consuming policy computation of \acro{ENS} with fast forward passes through the network.
Crucially, this network should learn a beneficial strategy that outperforms the greedy one-step policy, so that the cost of training and deploying the network outweighs one-step's speed and ease of use.
We now discuss our approach using reinforcement learning, specifically imitation learning, to effectively train one such network.

\subsection{Learning to search with imitation learning}

We start with the goal of training a neural network to learn to search using reinforcement learning (\acro{RL}), as the utility function in \cref{sec:as_formulation} can be naturally treated as a reward function, and each search run of $T$ iterations as belonging to a budget-constrained episodic Markov
decision process.
Given a search space defined by $\mathcal{X}$, the current state at iteration $t$ is given by $\mathcal{D}_t$, the data that we have collected thus far, while the unlabeled data points $\mathcal{X} \setminus \mathcal{D}_t$ make up the possible actions that can be taken.
Unlike many \acro{RL} settings, though, the size of the action space in a typical \acro{AS} problem makes it challenging for common \acro{RL} training algorithms.\footnote{Not to mention one of our main goals, scaling \acro{AS} to large search spaces.}
This is because many of these algorithms rely on thoroughly exploring the action space to learn about the value of each specific action in a given state, and as $n = | \mathcal{X} |$ grows larger, this task becomes increasingly more daunting.

Noting that we have access to \acro{ENS}, an expert policy with demonstrated superior performance throughout previous works, we forgo learning to search from scratch and seek to instead rely on \acro{ENS} for guidance.
This proves to be more feasible, as we can leverage imitation learning techniques in which we collect a data set of state and expert's action pairs $S = \{ s, \text{\acro{ENS}}(s) \}$ and train a neural network to learn this mapping.
Here, we wish our neural network to output the same decision generated by \acro{ENS} (i.e., which unlabeled point to query) given the current state (the observed data) of a search problem.
This is done by treating the goal of imitating the expert policy as a classification problem, where a data point is characterized by a given state $s$ of a search, and the corresponding label is the expert's decision $\text{\acro{ENS}}(s)$.
A neural network classifier is then trained to correctly classify $\text{\acro{ENS}}(s)$ as the desirable label among all possible actions, by minimizing the corresponding cross-entropy loss.

\setlength{\columnsep}{15pt}
\begin{wrapfigure}[17]{R}{0.46\textwidth}
\begin{minipage}{0.46\textwidth}
\vskip -23pt
\begin{algorithm}[H]
\setstretch{1.2}
\caption{DA\acro{GGER} for imitation learning}
\label{alg:dagger}
\begin{algorithmic}[1]
    \State {\bfseries inputs} number of training iterations $N$, expert policy $\pi_*$, problem generator $G$
    \State initialize $S \leftarrow \emptyset$
    \State initialize $\hat{\pi}_0$ randomly
    \For{$i = 1$ {\bfseries to} $N$}
        \State sample \acro{AS} problems $\mathcal{X} \sim G$
        \State roll out $\hat{\pi}_{i - 1}$ on $\mathcal{X}$ to obtain states $\{ s \}$
        \State assemble $S_i = \big\{ \big( s, \pi_*(s) \big) \big\}$
        \State aggregate $S \leftarrow S \cup S_i$
        \State train $\hat{\pi}_i$ on $S$ until convergence
    \EndFor
    \State {\bfseries returns} best $\hat{\pi}_i$ on validation
\end{algorithmic}
\end{algorithm}
\end{minipage}
\end{wrapfigure}

The training data $S$ for imitation learning can be assembled in several ways.
For example, we could run \acro{ENS} on training problems and record the states encountered and the decisions computed.
However, this leaves the possibility that as the trained network is deployed, it will arrive at a state very different from those seen during training, and thus output unreliable decisions.
We use DA\acro{GGER} \cite{ross2011reduction}, a well-established imitation learning technique, to address this problem.
DA\acro{GGER} is a meta-learning algorithm that iteratively rolls out the policy currently being trained (i.e., it uses the current policy to make decisions), collects the expert's actions on the encountered states, and appends this newly collected guidance to the training set to improve the policy being trained.
This iterative procedure allows the expert policy to be queried more strategically, targeting states the current policy network is likely to be in.
\cref{alg:dagger} summarizes this procedure.

\subsection{Constructing search problems for training}

To realize DA\acro{GGER}, we require access to a search problem ``generator''
that provides \acro{AS} problems in which we are free to roll out the network being trained and observe its performance.
One may consider directly using one's own real-world use case to train the policy network; however, running DA\acro{GGER} on real-life \acro{AS} problems might prove infeasible.
This is because DA\acro{GGER} is an iterative training loop that requires many training episodes to be played so that the collected training data $S$ could cover a wide range of behaviors of the expert to be imitated.
We cannot afford to dedicate many real search campaigns to this task, especially under our assumption of expensive labels.
Instead, we turn to synthetic problems generated in a way that is sufficiently diverse to present a wide range of scenarios under which we may observe \acro{ENS}'s behavior.
In addition to constructing these problems and running a policy currently being trained on them at little computational cost, we can limit the size of the problems so that \acro{ENS} can be queried efficiently.

When called, our data-generating process constructs a randomly generated set $\mathcal{X}$. 
We sample from a Gaussian process (\acro{GP}) \cite{rasmussen2006gaussian} at the locations in $\mathcal{X}$ to obtain a real-valued label for each $x \in \mathcal{X}$, which is then converted to a binary label by thresholding at a chosen quantile.
The generated search space and labels are returned as a training problem.
Although this procedure is quite simple and, in using a \acro{GP}, assumes a certain level of smoothness in the labels, we observe that the generated problems offer reasonable variety of structures with ``clumps'' of targets of variable number and size.
This variety successfully facilitates imitation learning, as later demonstrated by the empirical performance of our trained policy on real-world problems.
We include more details in \cref{sec:generate_problem}.

\subsection{Feature engineering \& implementation}
\label{sec:implementation}

The effectiveness of any training procedure in \acro{RL} crucially depends on the quality of the representation of a given state during a search.
To characterize a state in a way that aids learning, we use the following features to represent each unlabeled data point $x \in \mathcal{X} \setminus \mathcal{D}$ remaining in a search:
\setlist{nolistsep}
\begin{itemize}[noitemsep]
\item the posterior probability that the data point has a positive label $\Pr ( y = 1 \mid x, \mathcal{D} )$,
\item the remaining budget $\ell = T - t$,
\item the sum of posterior probabilities of the $(\ell - 1)$ unlabeled nearest neighbors of $x$:
\begin{equation}
\sum_{x' \in \text{\acro{NN}}(x, \, \ell - 1)} \Pr ( y' = 1 \mid x', \mathcal{D} ),
\end{equation}
where $\text{\acro{NN}}(x, k)$ denotes the set of $k$ unlabeled nearest neighbors of a given $x \in \mathcal{X}$, and
\item the sum of similarities between $x$ and its $(\ell - 1)$ unlabeled nearest neighbors:
\begin{equation}
\sum_{x' \in \text{\acro{NN}}(x, \, \ell - 1)} s(x, x'),
\end{equation}
where $s(x, x') \in [0, 1]$ denotes the similarity between two given points $x, x' \in \mathcal{X}$. 
\end{itemize}

Which similarity function $s$ to use to compute the nearest neighbors of each point and the corresponding similarity values depends on the application.
We use the radial basis function kernel $s(x, x') = \exp \big( -\tfrac{\| x - x' \|^2}{2 \, \lambda^2} \big)$  during training and upon deployment for appropriate tasks in \cref{sec:experiments}, but this can be replaced with another function more applicable to a given domain.  

We specifically design the third and fourth features to relate the value of an unlabeled point we may query to the characteristics of its nearest neighbors.
Intuitively, a point whose neighbors are likely targets is a promising candidate, as it indicates a region that could yield many hits.
On the other hand, points that are close to its neighbors (e.g., cluster centers) could also prove beneficial to query, as they help the policy explore the space effectively.
Further, the number of nearest neighbors to include in these computations is set to match the length of the horizon, allowing these features to dynamically adjust to our remaining budget.
Overall, the four features make up the feature vector of each candidate point, and concatenating all feature vectors gives the state representation of a given search iteration.
We also note that our state representation is task-agnostic and applicable across \acro{AS} problems of varying structures and sizes, which is crucial for training our policy on the different problems we generate, as well as for when we deploy our trained policy on unseen problems.

Here, finding the nearest neighbors of each point may prove challenging under large spaces.
We leverage the state-of-the-art similarity search library F\acro{AISS} \cite{johnson2019billion} to perform efficient approximate nearest neighbor search when exact search is prohibitive.\footnote{We set the number of clusters into which the search space is split when performing approximate neighbor search at $\lfloor 4 \sqrt{n} \rfloor$, following \url{https://github.com/facebookresearch/faiss/issues/112}.}
F\acro{AISS} allows us to significantly accelerate this step, completing, for example, the neighbor search for our largest problem in \cref{sec:experiments} of 6.7 million points in roughly one hour.
Once this neighbor search is done before the actual search campaign, the time complexity of constructing the features above at each search iteration is $O(n)$.

We run \cref{alg:dagger} for $N = 50$ iterations, each consisting of 3 training problems.
At the end of each iteration, we train a small policy network with 5 fully connected hidden layers (with 8, 16, 32, 16, and 8 neurons, respectively) and \acro{R}e\acro{LU} activation functions using minibatch gradient descent with Adam optimizer \cite{kingma2015adam}.
The trained policy is  then evaluated on a fixed set of 3 unseen validation problems.
We set the labeling budget $T = 100$ across all generated problems.

\subsection{Demonstration of learned search strategy}

\begin{figure}
\centering
\includegraphics[width=\linewidth]{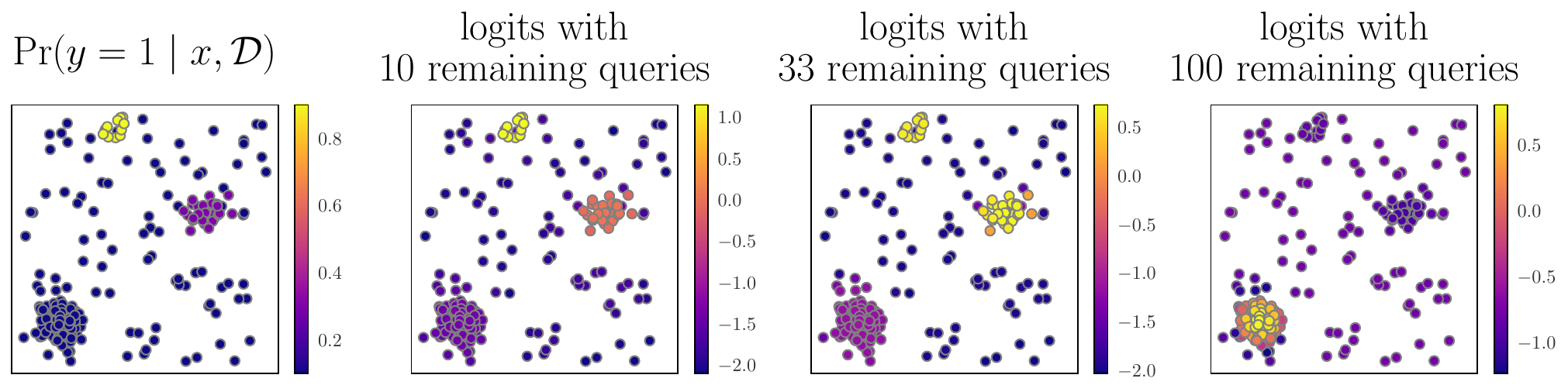}
\vskip -5pt
\caption{
Demonstration of our trained policy's budget-awareness with a toy example.
\textbf{Left panel}: the probability that a point is a target.
\textbf{Remaining panels}: computed logits and the point selected to be the next query under different labeling budgets.
Our policy appropriately balances between exploitation under a small labeling budget and strategic exploration if the budget is large.
}
\vskip -10pt
\label{fig:budget_aware_toy}
\end{figure}

Before discussing our experiment results, we briefly demonstrate the learned behavior of our trained policy network using an illustrative toy example visualized in \cref{fig:budget_aware_toy}.
The search space $\mathcal{X}$ consists of uniformly sampled points as well as 3 distinct clusters of different sizes.
The left panel shows $\Pr ( y = 1 \mid x, \mathcal{D} )$, the probability that each point is a target, specifically set so that:
\setlist{nolistsep}
\begin{itemize}[noitemsep]
\item for each of the 100 uniformly sampled points, $\Pr ( y = 1 \mid x, \mathcal{D} ) = 0.1$,
\item for each point in the small cluster of size 10 at the top, $\Pr ( y = 1 \mid x, \mathcal{D} ) = 0.9$,
\item for each point in the medium cluster of size 30 on the right, $\Pr ( y = 1 \mid x, \mathcal{D} ) = 0.3$, and
\item for each point in the large cluster of size 100 on the bottom left, $\Pr ( y = 1 \mid x, \mathcal{D} ) = 0.1$.
\end{itemize}
This problem structure presents an interesting choice between exploiting the small cluster of very likely targets and exploring larger clusters that contain less likely targets.
A good policy should select exploitation if the remaining budget is small and choose to further explore otherwise.
The remaining panels in \cref{fig:budget_aware_toy}, which visualize the logits computed by our trained policy under different remaining budgets $\ell \in \{ 10, 33, 100 \}$, show that this is exactly the case: the policy targets the cluster of likely targets when the budget is small, and moves to larger clusters as the budget increases.
Further, when exploring, the policy appropriately favors cluster centers, which offer more information about the space.
This balance between exploitation and strategic exploration our trained policy exhibits indicates that the policy has learned a meaningful search behavior from \acro{ENS}, which translates into good empirical performance, as later shown in \cref{sec:experiments}.

\section{Related work}

\textbf{Active search.}
We continue the line of research on active search (\acro{AS}) \cite{garnett2012bayesian,coleman2022similarity}, which previous works have also referred to as active learning and adaptive sampling for discovery \cite{warmuth2002active,warmuth2003active,xu2022adaptive} or active covering \cite{jiang2021active}.
\citet{garnett2012bayesian} studied the Bayesian optimal policy, and \citet{jiang2017efficient} proposed \acro{ENS} as a budget-aware approximation demonstrating impressive empirical success.
\acro{ENS} has since then been adopted under various settings, including batch \cite{jiang2018efficient}, cost-aware \cite{warmuth2002active,warmuth2003active,jiang2019cost}, multifidelity \cite{nguyen2021nonmyopic}, and diversity-aware \acro{AS} \cite{nguyen2023nonmyopic}.
We propose to amortize policy computation of \acro{ENS} using imitation learning, scaling nonmyopic search to large data sets.

\textbf{Amortization via neural networks.}
Using neural networks to amortize expensive computations has seen increasing interests from the machine learning community.
Of note is the work of \citet{foster2021deep}, who tackled amortizing maximizing expected information gain \cite{mackay1992evidence,mackay1992information} for Bayesian experimental design (\acro{BED}) \cite{lindley1956measure} with a design network trained on a specialized loss function, and was a major inspiration for our work.
Subsequent works \cite{blau2022optimizing,shen2023bayesian} have studied \acro{BED} under other settings such as those with discrete action spaces.
\citet{liu2020task} tackled amortizing Gaussian process (\acro{GP}) inference by training a transformer-based network as a regression model to predict optimal hyperparameters of \acro{GP}s with stationary kernels; \citet{bitzer2023amortized} later extended the approach to more general kernel structures.
\citet{andrychowicz2016learning}, on the other hand, learned an optimization policy with a recurrent neural network that predicts the next update to the parameters to be optimized based on query history; the policy network was shown to outperform many generic gradient-based optimizers.
Also related is the work of \citet{konyushkova2017learning}, where a regressor was trained to predict the value of querying a given unlabeled data point for active learning.

\textbf{Reinforcement learning.}
Reinforcement learning (\acro{RL}) has proven a useful tool for learning effective strategies for planning tasks similar to \acro{AS}.
Examples include active learning policies for named entity recognition \cite{fang2017learning,liu2018learning}, neural machine translation \cite{liu2018learning_neural}, and active learning on graphs \cite{hu2020graph}.
\citet{igoe2021multi} were interested in path planning for drones, framed as a specialized \acro{AS} setting with linear models and many agents.
\citet{sarkar2023partially} and \citet{sarkar2024visual} studied the problem of visual \acro{AS}, a realization of \acro{AS} on images for geospatial exploration.
Overall, the methodologies in these works rely on being able to generate many training episodes to learn an effective \acro{RL} policy from scratch, which cannot be realized in our setting.
Having access to \acro{ENS}, we instead leverage imitation learning to learn to search from this expert on synthetically generated search problems.
When deployed, our trained policy can be applied to a diverse set of use cases, as demonstrated in the next section.

\renewcommand{\arraystretch}{1.2}
\begin{table}[t]
\centering
\caption{
Average number of targets found and standard errors by each search policy across 10 repeats of all instances of a given task.
Settings that are computationally prohibitive for a given policy are left blank.
The best policy in each setting is highlighted \textbf{bold}; policies not significantly worse than the best (according to a two-sided paired $t$-test with a significance level of $\alpha = 0.05$) are in \emph{\color{blue}{blue italics}}.
}
\resizebox{\textwidth}{!}{\begin{tabular}{c >{\centering\arraybackslash}m{2.3cm} >{\centering\arraybackslash}m{2.3cm} >{\centering\arraybackslash}m{2.3cm} >{\centering\arraybackslash}m{2.3cm} >{\centering\arraybackslash}m{2.3cm} >{\centering\arraybackslash}m{2.3cm}}
\toprule
& Disease hotspots & \acro{F}ashion-\acro{MNIST} & Bulk metal glass & Drug discovery (small) & GuacaMol & Drug discovery (large) \\
\midrule
\acro{ETC}$(m = 10)$ & \result{50.50}{4.44} & \result{47.79}{3.89} & \result{81.70}{3.66} & \result{75.46}{2.64} & \result{10.37}{1.80} & \resultblueit{33.97}{4.14} \\
\acro{ETC}$(m = 20)$ & \result{52.25}{3.47} & \result{42.97}{3.47} & \result{73.40}{3.67} & \result{68.46}{2.37} & \result{9.42}{1.61} & \result{30.42}{3.70} \\
\acro{ETC}$(m = 30)$ & \result{48.75}{3.12} & \result{38.16}{3.03} & \result{65.70}{3.17} & \result{60.76}{2.11} & \result{8.56}{1.45} & \result{26.59}{3.24} \\
\midrule
\acro{UCB}$(\beta = 0.1)$ & \result{48.20}{4.21} & \result{51.66}{4.24} & \resultblueit{88.80}{4.00} & \result{80.74}{2.86} & \result{11.21}{1.93} & \resultblueit{36.94}{4.49} \\
\acro{UCB}$(\beta = 0.3)$ & \result{48.15}{4.20} & \result{51.66}{4.24} & \resultblueit{88.80}{4.00} & \result{80.74}{2.86} & \result{11.21}{1.93} & \resultblueit{36.94}{4.49} \\
\acro{UCB}$(\beta = 1)$ & \result{45.48}{3.52} & \result{51.10}{4.18} & \result{81.80}{3.15} & \result{75.14}{2.59} & \result{11.21}{1.93} & \resultblueit{36.94}{4.49} \\
\midrule
one-step & \result{48.20}{4.21} & \result{51.66}{4.24} & \resultblueit{88.80}{4.00} & \result{80.74}{2.86} & \result{11.21}{1.93} & \resultblueit{36.94}{4.49} \\
\acro{IDS} & \result{48.83}{4.08} & \result{51.66}{4.24} & --- & --- & --- & --- \\
\acro{ENS} & \resultbold{57.67}{3.22} & \resultbold{88.15}{1.52} & \resultbold{91.10}{3.48} & \resultbold{86.82}{2.41} & --- & --- \\
\midrule
\acro{ANS} (ours) & \resultblueit{57.25}{3.58} & \result{85.72}{1.69} & \resultblueit{89.90}{4.20} & \resultblueit{85.29}{2.47} & \resultbold{15.51}{2.36} & \resultbold{39.83}{3.76} \\
\bottomrule
\end{tabular}}
\vskip -10pt
\label{tab:utility}
\end{table}

\section{Experiments}
\label{sec:experiments}

We tested our policy network, which we call \emph{amortized nonmyopic search}, or \acro{ANS}, and a number of baselines on search problems spanning a wide range of applications.
For each problem included, we run each policy 10 times from the same set of initial data $\mathcal{D}_0$ that contains one target and one non-target, both randomly sampled.
Each of these runs has a labeling budget of $T = 100$.

\textbf{Baselines.}
We compare our method against the expert policy \acro{ENS} which ours was trained to mimic, as well as the one-step policy that greedily queries the most likely target discussed in \cref{sec:as_formulation}.
We also implement a number of baseline policies from the literature.
The first is a family of upper confidence bound (\acro{UCB}) policies \cite{auer2000using,carpentier2011upper} that rank candidates by the following score: $p + \beta \sqrt{p (1 - p)}$, where $p = \Pr ( y = 1 \mid x, \mathcal{D} )$ is the posterior probability that a given candidate is a target, and $\beta$ is the tradeoff parameter balancing exploitation (favoring large $p$) and exploration (favoring large uncertainty in the label, as measured by $\sqrt{p (1 - p)}$).
Here, we have $\beta$ take on values from $\{ 0.1, 0.3, 1 \}$.
Another baseline is from \citet{jiang2021active}, who proposed a simple explore-then-commit (\acro{ETC}) scheme that uniformly samples the space for $m$ iterations and then switches to the greedy sampling strategy of one-step for the remaining of the search.
We run this \acro{ETC} policy with $m \in \{ 10, 20, 30 \}$.
Finally, we include the policy by \citet{xu2022adaptive}, which uses the information-directed sampling (\acro{IDS}) heuristic \cite{russo2014learning,russo2018learning} that scores each candidate query $x$ by the ratio between (1) information about the labels of the current $\ell$ most likely targets ($\ell = T - t$ is the length of the remaining horizon), gained by querying $x$ and (2) the expected instant regret from querying $x$.
We note that \citet{xu2022adaptive} proposed this policy under specialized \acro{AS} settings that allow information gain to be efficiently computed; they also only considered problems with fewer than 1000 points.
For our experiments, we can only apply \acro{IDS} to relatively small spaces where the policy is computationally feasible.

\begin{figure}
\centering
\includegraphics[width=0.49\linewidth]{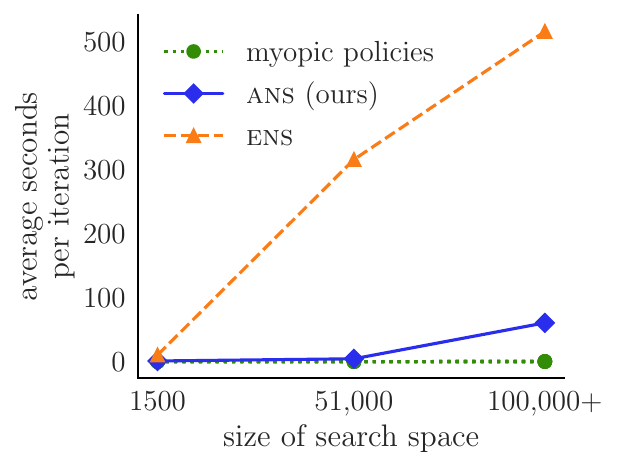}
\hfill
\includegraphics[width=0.49\linewidth]{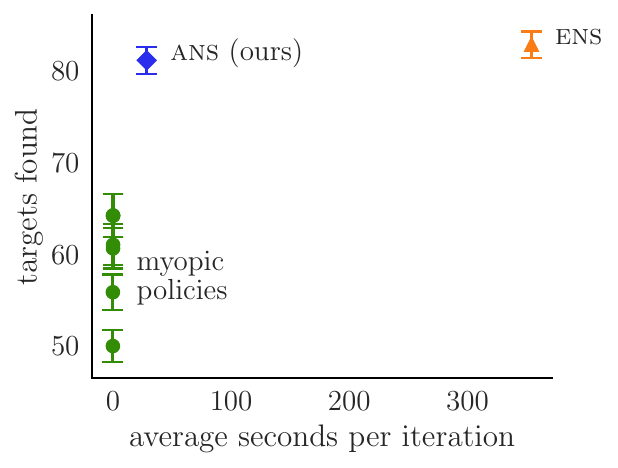}
\vskip -10pt
\caption{
The time taken per iteration by different active search policies in the small- and medium-scale experiments.
\textbf{Left}: average number of seconds per iteration with respect to the size of the search space.
\textbf{Right}: average number of targets found and standard errors vs.\ time per iteration.
}
\vskip -10pt
\label{fig:time}
\end{figure}

\textbf{Search problems.}
We now discuss the search problems making up our experiments.
The first was posed by \citet{andrade2020finding}, who sought to identify disease hotspots within a region of interest.
The provided data sets correspond to 4 distinct \acro{AS} problems of finding locations with a high prevalence of schistosomiasis in C\^{o}te d'Ivoire and Malawi and of lymphatic filariasis in Haiti and the Philippines.
Each problem consists of 1500 points, with targets accounting for 10\%--34\% of the space.
For our second task, following \citet{nguyen2023nonmyopic}, we simulate product recommendation problems using the \acro{F}ashion-\acro{MNIST} data \cite{xiao2017fashion}, which contains 70\,000 images classified into 10 classes of clothing articles.
We first randomly select 3 out of the 10 classes as products a user is interested in (i.e., our search targets).
We further sub-sample these 3 classes uniformly at random to increase the difficulty of the problem; the resulting prevalence rate of the targets is roughly 6\%.
We repeat this process 10 times to generate 10 search problems with this data.

Borrowing from previous works \cite{jiang2017efficient,jiang2018efficient}, we use a data set from the materials science literature \cite{kawazoe1997nonequilibrium,ward2016general} containing 106\,810 alloys, of which 4275 can form bulk metallic glasses with high toughness and wear resistance and are our search targets.
Another application comes from drug discovery, where we aim to identify ``active compounds'', chemical compounds that bind with a targeted protein.
\citet{garnett2015introducing} assembled a suite of such drug discovery problems, each of which consists of the active compounds for a specific protein from the \acro{B}inding\acro{DB} database \cite{liu2007bindingdb}, and 100\,000 molecules sampled from the \acro{ZINC} database \cite{sterling2015zinc} that act as the negative pool. 
Our experiments include the first 10 problems where on average the active compounds make up 0.5\% of the search space.

Finally, to demonstrate the ability to perform search on large spaces achieved by our method \acro{ANS}, we consider two large-scale, challenging drug discovery tasks.
The first employs the GuacaMol database of over 1.5 million drug-like molecules that were specifically curated for drug discovery benchmarking tasks involving machine learning \cite{brown2019guacamol}.
In addition to these molecules, the database offers a family of objective functions to measure the molecules' quality using a variety of criteria.
We use each objective function provided to define a search problem as follows.
We first randomly sample a set of 1000 molecules which we fully label using the objective functions.
We then define the search targets as those of the remaining unlabeled molecules whose scores exceed the 99-th percentile of the labeled set; in other words, the goal of our search is the top 1\% molecules.
In total, we assemble 9 such \acro{AS} problems with GuacaMol.
For our second task, we follow the procedure by \citet{garnett2015introducing} described above with the \acro{B}inding\acro{DB} and \acro{ZINC} databases, this time expanding the negative pool to all drug-like molecules in \acro{ZINC} \cite{tingle2023zinc}.
This results in a search space of 6.7 million candidates, of which 0.03\% are the active compounds we aim to search for.

\textbf{Discussions.} \cref{tab:utility} reports the performance of the search policies -- measured in the number of targets discovered -- in each of these tasks, where settings that are computationally prohibitive for a given policy are left blank.
From these results, we observe a clear trend: the state-of-the-art policy \acro{ENS} consistently achieves the best performance under all settings that it could feasibly run, while our trained policy network \acro{ANS} closely follows \acro{ENS}, sometimes outperforming the other baselines by a large margin.
Our method also yields the best result in large-scale problems, demonstrating its usefulness in large search spaces.
Among the baselines, we note the difficulty in setting the number of exploration rounds $m$ for \acro{ETC}, since no value of $m$ performs the best across all settings.
Results from \acro{UCB} policies, on the other hand, indicate that prioritizing exploitation (setting the parameter $\beta$ to a small value) is beneficial, a trend also observed in previous work \cite{jiang2017efficient}.

To illustrate the tradeoff between performance and speed achieved by \acro{ANS}, the left panel of \cref{fig:time} shows the average time taken by \acro{ANS}, \acro{ENS}, and myopic baselines per iteration as a function of the size of the data, while the right panel shows the number of discoveries by each policy vs.\ the same average time per iteration.
These plots do not include the results from the large-scale problems so that the comparison with \acro{ENS} is fair, or \acro{IDS} which is slower than \acro{ENS} but does not perform as well.
We see that \acro{ANS} finds almost as many targets as \acro{ENS} but is much more computationally lightweight.
We thus have established a new point on the Pareto frontier of the performance vs.\ speed tradeoff with our search policy.
In the 6.7 million-point drug discovery problems, \acro{ANS} on average takes 36.94 $\pm$ 0.15 minutes per iteration, which we deem entirely acceptable given the boost in performance compared to faster but myopic baselines, the fact that the time cost of labeling is typically much higher, and \acro{ENS}, in comparison, is estimated to take roughly 10 hours per iteration on the same scale.

\begin{figure}
\centering
\includegraphics[width=\linewidth]{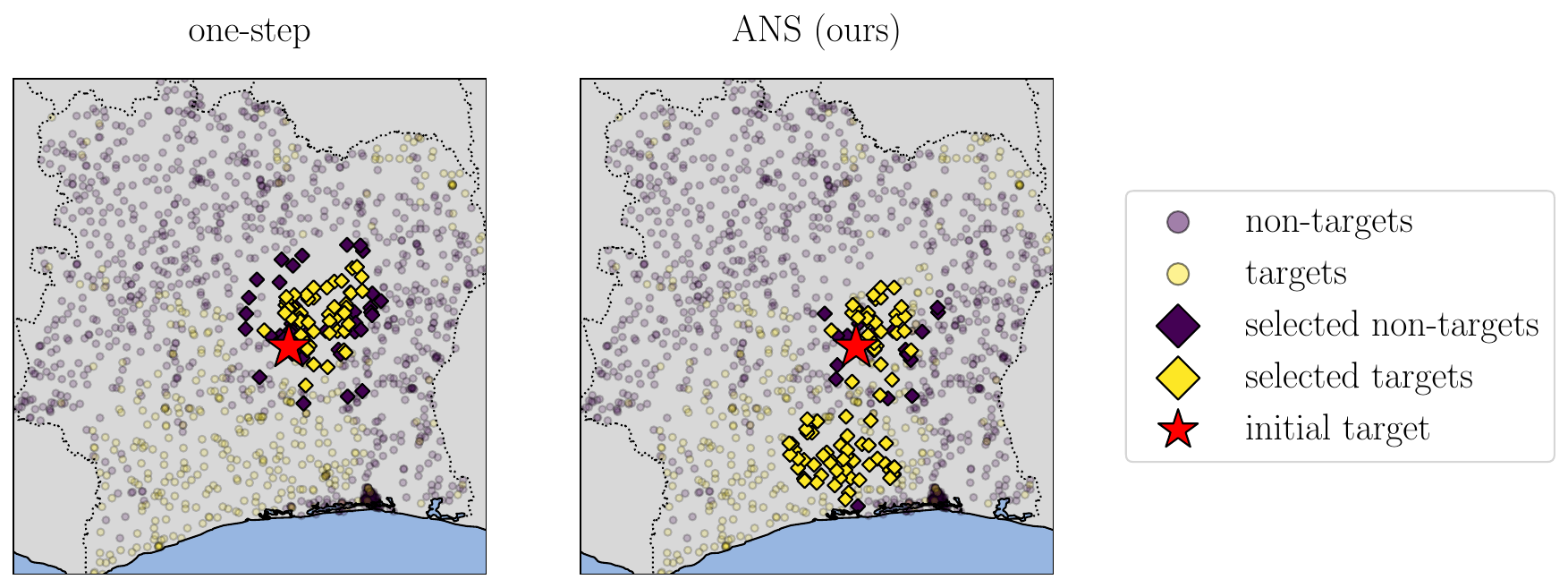}
\vskip -5pt
\caption{
Locations in C\^{o}te d'Ivoire selected by the one-step policy and by ours in an illustrative run with the disease hotspot data, where our policy discovers a larger target cluster.
}
\vskip -10pt
\label{fig:hotspot_civ_with_star_v2}
\end{figure}

\setlength{\columnsep}{15pt}
\begin{wrapfigure}[20]{r}{0.4\textwidth}
\centering
\vskip -13pt
\includegraphics[width=0.4\textwidth]{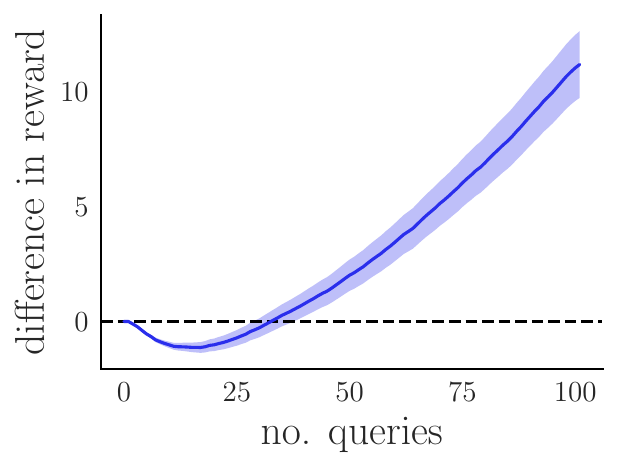}
\vskip -5pt
\caption{
The average difference in cumulative reward and standard errors between our policy and one-step.
Our policy spends its initial budget exploring the space and finds fewer targets in the beginning but smoothly switches to more exploitative queries and outperforms one-step at the end.
}
\label{fig:reward_diff}
\end{wrapfigure}

As a demonstration of our policy's strategic explorative behavior learned from \acro{ENS}, \cref{fig:hotspot_civ_with_star_v2} shows the result of an illustrative run by the one-step policy vs.\ \acro{ANS} from the problem of finding schistosomiasis hotspots in C\^{o}te d'Ivoire \cite{andrade2020finding}.
We note that the queries made by one-step are localized within the center region 
containing the target in the initial data $\mathcal{D}_0$, while \acro{ANS} is able to discover a larger cluster of targets to the south.
Further, \cref{fig:reward_diff} visualizes the cumulative difference in utility between \acro{ANS} and one-step across all experimental settings. 
Here, \acro{ANS} initially finds fewer targets than one-step, as the former tends to dedicate its queries to exploration of the space when the remaining budget is large; however, as the search progresses, \acro{ANS} smoothly transitions to more exploitative queries and ultimately outperforms the greedy policy.
The same pattern of behavior has been observed from \acro{ENS} in previous works \cite{jiang2017efficient,nguyen2021nonmyopic,nguyen2023nonmyopic}.

We defer further analyses to \cref{sec:more_experiments}, which includes an ablation study showing the importance of each feature we engineer for our policy network in \cref{sec:implementation}, as well as the benefit of the DA\acro{GGER} loop compared to (1) imitation learning without iteratively generating more training data and (2) learning to search from scratch without \acro{ENS}.
We also examine the stability between different training runs giving different policy networks, and observe that these policies yield comparable results.
Finally, we investigate the effectiveness of various schemes to further improve search performance under settings where repeated searches are conducted within the same space.

\section{Conclusion}

We propose an imitation learning-based method to scale nonmyopic active search to large search spaces, enabling real-time decision-making and efficient exploration of massive databases common in product recommendation and drug discovery beyond myopic/greedy strategies.
Extensive experiments showcase the usefulness of our policy, which mimics the state-of-the-art policy \acro{ENS} while being significantly cheaper to run.
Future directions include deriving a more effective reinforcement learning strategy to train our policy network, potentially outperforming \acro{ENS}, as well as extending to other active search settings such as batch \cite{jiang2017efficient} and diversity-aware search \cite{nguyen2023nonmyopic,nguyen2024quality}.

\bibliography{main}
\bibliographystyle{plainnat}

\newpage

\appendix

\section{Generation of training search problems}
\label{sec:generate_problem}

\setlength{\columnsep}{15pt}
\begin{wrapfigure}[18]{r}{0.3\textwidth}
\centering
\vskip -15pt
\includegraphics[width=0.3\textwidth]{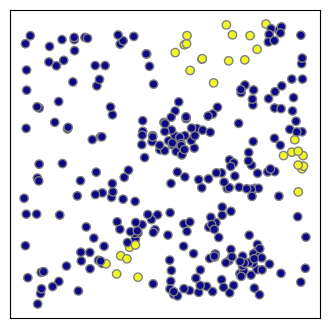}
\vskip -5pt
\caption{
An example two-dimensional problem generated by \cref{alg:generate_problem}, where bright and dark points indicate targets and non-targets, respectively.
The search space includes both clusters and more widely dispersed points.
}
\label{fig:problem_2d}
\end{wrapfigure}

We now discuss our procedure of generating active search problems to train the policy network in DA\acro{GGER}, summarized in \cref{alg:generate_problem}, where $U \{ m, n \}$ denotes a discrete uniform distribution of the integers between $m$ and $n$ (inclusive), while $U[a, b]$ refers to a continuous uniform distribution between $a$ and $b$.

The search space $\mathcal{X}$ of each generated problem exists in a $d$-dimensional space, where $d$ is a random integer between $2$ and $10$.
Once $d$ is determined, we sample $100 d$ points uniformly from the $d$-dimensional unit hypercube.
This set of uniform points is combined with $n_\text{cluster}$ clusters, where $n_\text{cluster}$ is a random integer between $10$ and $10d$.
To generate each cluster, we first sample another random integer between $10$ and $10d$, denoted as $m$, to determine the size of the cluster.
We then draw $m$ points from an isotropic Gaussian distribution with the mean vector $\boldsymbol{\mu}$ randomly sampled within the unit hypercube and the diagonal covariance matrix $\sigma^2 I_d$, where $\sigma$ is drawn from $U[0.1, 0.1 d]$.
Here, $\sigma$, which determines the spread of a given cluster, is constrained to be between $0.1$ and $0.1 d$ (relatively small numbers) to ensure that the points within this cluster are indeed close to one another.

Again, the union of the uniform points and the clusters make up the entire search space $\mathcal{X}$.
We then draw a sample from a Gaussian process (\acro{GP}) at the points in $\mathcal{X}$.
This \acro{GP} is equipped with a zero mean function and a radial basis function kernel whose length scale $\ell$ scales linearly with the dimensionality of the space $d$ by a factor of $0.05$.
This \acro{GP} sample yields a vector $\mathbf{f}$ of real-valued numbers.
We then sample uniformly between $0.01$ and $0.2$ for a prevalence rate $p$, which determines the proportion of $\mathcal{X}$ corresponds to the targets.
As such, we compute the binary labels $\mathbf{y}$ by thresholding $\mathbf{f}$ at the $100(1 - p)$-th quantile of the values in $\mathbf{f}$.
We keep the prevalence rate $p$ below 20\% to ensure that the targets are sufficiently rare.
The tuple $(\mathcal{X}, \mathbf{y})$ is finally returned.
\cref{fig:problem_2d} shows an example of one such generated problem in two dimensions, showing a search space with a considerably complex structure with multiple clusters and groups of targets.

\section{Data sets}
\label{sec:data}

We now describe the data sets used in our experiments in \cref{sec:experiments}.
These data sets are curated from authors of respective publications respecting their licenses, as detailed below.
No identifiable information or offensive content is included in the data.
\setlist{nolistsep}
\begin{itemize}[noitemsep]
\item We downloaded the disease hotspot data set from the GitHub repository provided in \citet{andrade2020finding}.
We computed the nearest neighbors of each data point (a location within one of the four countries included) using its coordinates (longitude and latitude).
\item The \acro{F}ashion-\acro{MNIST} data set is published by \citet{xiao2017fashion}.
We used \acro{UMAP} \citep{mcinnes2018umap} to produce a two-dimensional embedding of the images and compute the nearest neighbors on this embedding.
\item We obtained the bulk metal glass data from \citet{jiang2017efficient}, who, following \citet{ward2016general}, represented each data point with various physical attributes that were found to be informative in predicting glass-forming ability.
Each feature is subsequently scaled to range between $0$ and $1$.
The nearest neighbor search is performed on these features.
\item The data for the first set of drug discovery problems were also obtained from \citet{jiang2017efficient}, where the Morgan fingerprints \citep{rogers2010extended} were used as the feature vectors and the Tanimoto coefficient \citep{willett1998chemical} as the measure of similarity.
\item The molecules in the GuacaMol data are included in the publication of \citet{brown2019guacamol}, while those in our large drug discovery tasks were downloaded from the \acro{ZINC}-22 database \cite{tingle2023zinc}.
For each of these data sets, we used the state-of-the-art transformer-based molecular variational autoencoder trained in \citet{maus2022local} to generate a 256-dimensional embedding of the molecules.
The approximate nearest neighbor search described in \cref{sec:implementation} was conducted on this embedding.
\end{itemize}

\begin{algorithm}[t]
\setstretch{1.2}
\caption{Generate synthetic search problems}
\label{alg:generate_problem}
\begin{algorithmic}[1]
    \State $d \sim U \{ 2, 10 \}$ \Comment{sample dimensionality of search space}
    \State $\mathcal{X} \leftarrow \big\{ x_j: x_j \sim U [0, 1] \big\}_{j = 1}^{100 d}$ \Comment{sample uniform points}
    \State $n_\text{cluster} \sim U \{ 10, 10d \}$ \Comment{sample number of clusters}
    \For{$i = 1$ {\bfseries to } $n_\text{cluster}$}
        \State $m \sim U \{ 10, 10d \}$ \Comment{sample cluster size}
        \State $\boldsymbol{\mu} = [ \mu_j ]_{j = 1}^d$, where $\mu_j \sim U [0, 1]$ \Comment{sample cluster center}
        \State $\sigma \sim U [0.1, 0.1 d]$ \Comment{sample spread of cluster}
        \State $\mathcal{X} \leftarrow \mathcal{X} \cup \big\{ x_j: x_j \sim \mathcal{N}(\boldsymbol{\mu}, \sigma^2 I_d) \big\}_{j = 1}^m$ \Comment{use an isotropic Gaussian distribution}
    \EndFor
    \State $\mathbf{f} \sim \mathcal{N}(\mathcal{X}; \mathbf{0}, \mathbf{\Sigma})$, \newline where $\mathbf{\Sigma} = K(\mathcal{X}, \mathcal{X})$ and $K(\mathbf{x}_1, \mathbf{x}_2) = \exp \left( - \tfrac{\| \mathbf{x}_1 - \mathbf{x}_2 \|^2}{2 \ell^2} \right)$ with length scale $\ell = 0.05 d$
    \State $p \sim U[0.01, 0.2]$ \Comment{sample target prevalence}
    \State $\mathbf{y} = \mathbb{I} \big[ \mathbf{f} > 100 (1 - p)$\text{-th quantile in } $\mathbf{f} \big]$ \Comment{threshold to construct binary labels}
    \State {\bfseries returns} $(\mathcal{X}, \mathbf{y})$
\end{algorithmic}
\end{algorithm}

\section{Further details on experiments}
\label{sec:more_experiments}

Experiments were performed on a small cluster built from commodity hardware comprising approximately 200 Intel Xeon \acro{CPU} cores, each with approximately 10 \acro{GB} of \acro{RAM}.
All compute amounted to roughly 15\,000 \acro{CPU} hours, including preliminary experiments not discussed in the paper, training the policy network, and all experiments for evaluation discussed in \cref{sec:experiments} and here.

\renewcommand{\arraystretch}{1.2}
\begin{table}[t]
\centering
\caption{
Average number of targets found and standard errors by each ablated policy across 100 product recommendation tasks with \acro{F}ashion\acro{MNIST}.
The best policy is highlighted \textbf{bold}.
}
\resizebox{\textwidth}{!}{\begin{tabular}{>{\centering\arraybackslash}m{2.3cm} >{\centering\arraybackslash}m{2.3cm} >{\centering\arraybackslash}m{2.3cm} >{\centering\arraybackslash}m{2.3cm} >{\centering\arraybackslash}m{2.3cm} >{\centering\arraybackslash}m{2.3cm} >{\centering\arraybackslash}m{2.3cm}}
\toprule
without target probability & without remaining budget & without neighbor probability sum & without neighbor similarity sum & imitation learning without DA\acro{GGER} & \acro{REINFORCE} without imitation learning & \acro{ANS} (ours) \\
\midrule
\result{66.93}{1.51} & \result{80.11}{1.80} & \result{73.35}{1.61} & \result{63.29}{3.35} & \result{59.66}{3.80} & \result{75.15}{1.52} & \resultbold{85.72}{1.69} \\
\bottomrule
\end{tabular}}
\label{tab:utility_ablation}
\end{table}

\setlength{\columnsep}{15pt}
\begin{wrapfigure}[21]{r}{0.5\textwidth}
\centering
\vskip -15pt
\includegraphics[width=0.5\textwidth]{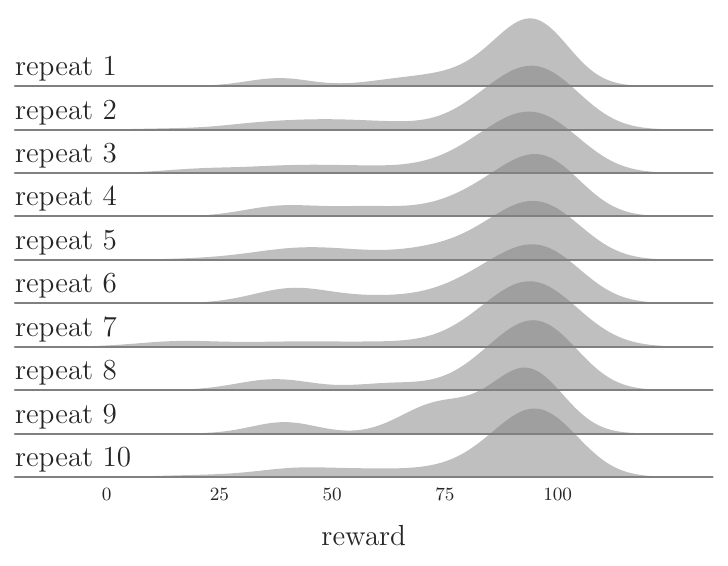}
\vskip -10pt
\caption{
Distributions of the number of targets found across 100 product recommendation tasks with \acro{F}ashion\acro{MNIST} by 10 policy networks trained with different initial random seeds.
The distributions are comparable, indicating that the trained policy networks behave similarly.
}
\label{fig:stability_fasion}
\end{wrapfigure}

\textbf{Ablation study.}
We use the 10 product recommendation tasks from the \acro{F}ashion\acro{MNIST} data to quantify the value of various components of our framework.
First, we trained four additional policy networks, each learning from \acro{ENS} without one of the four features discussed in \cref{sec:implementation}.
We also trained another network using imitation learning but without DA\acro{GGER}'s iterative procedure: we ran the expert policy \acro{ENS} on $3 \times 50 = 150$ generated search problems (the same number of problems generated to train the policy examined in the main text), kept track of the encountered states and selected actions, and used these data to train the new network until convergence only once.
Finally, we trained a policy network without imitation learning using the \acro{REINFORCE} policy gradient algorithm \cite{sutton1999policy}.
The performance of these policies, along with that of our main policy \acro{ANS} as a reference, is shown in \cref{tab:utility_ablation}.
We see that by removing any component of our imitation learning procedure, we incur a considerable decrease in search performance, which demonstrates the importance of each of these components.

\textbf{Training stability.}
We rerun our training procedure with DA\acro{GGER} for 10 times using different random seeds and evaluate the trained policy networks using the experiments with the \acro{F}ashion\acro{MNIST} data.
Each row of \cref{fig:stability_fasion} shows the distribution of the number of targets found by each of these 10 policy networks across the 100 search problems.
We observe that the variation across these 10 distributions is quite small, especially compared to the variation across different search runs by the same policy network.
This shows that our training procedure is stable, resulting in policy networks that behave similarly under different random seeds.

\textbf{Refinement under repeated search.}
In many settings that \acro{AS} targets, multiple search campaigns may be conducted within the same search space.
For example, as in our drug discovery experiments in \cref{sec:experiments}, a scientist may explore a molecular database to identify candidates with different desirable properties.
As the search for a given property concludes, the next search stays within the same database but now targets a different property.
In these situations, we may reasonably seek to refine our search strategy throughout these episodes using the results we observe, so that our search policy could improve using its past experiences.
We identify two approaches to such refinement:
\setlist{nolistsep}
\begin{itemize}[noitemsep]
\item If a neural network is used as the search policy, it can be updated by a policy gradient algorithm such as \acro{REINFORCE} \cite{sutton1999policy} after each episode.
\item If a deep autoencoder (\acro{DAE}) is used to produce a representation of the search candidates (on which the nearest neighbor search described in \cref{sec:implementation} is conducted), the autoencoder can be updated with a semisupervised loss \cite{kingma2014semi} that accounts for the labels it iteratively uncovers throughout the search.
\end{itemize}

To investigate the effects of each of these approaches on the search performance of our policy trained with imitation learning and examined in the main text, we engineer another version of the \acro{F}ashion\acro{MNIST} data set \cite{xiao2017fashion} that simulates a setting of repeated search.
We first randomly choose 5 out of 10 classes in the data set to act as possible target sets throughout the repeated searches.
These selected classes are then sub-sampled uniformly at random so that there are only 1000 data points per class; this yields a data set of 40\,000 points in total.
We then use a variational autoencoder \cite{kingma2014auto} to learn a two-dimensional representation of these 40\,000 candidates.

We allow 100 search episodes within this database, where in each episode, 1 of the chosen 5 classes is randomly selected as the target class.
To implement the second approach to search refinement, we train a variational Gaussian process classifier on the observed data $\mathcal{D}$ and use the corresponding evidence lower bound (\acro{ELBO}) to make up the supervised component of the joint loss of the semisupervised model.
While we update the search policy using the \acro{REINFORCE} loss at the end of each episode, an update to the semisupervised \acro{VAE} is performed for every 20 iterations within one episode.

\begin{figure}
\centering
\includegraphics[width=\linewidth]{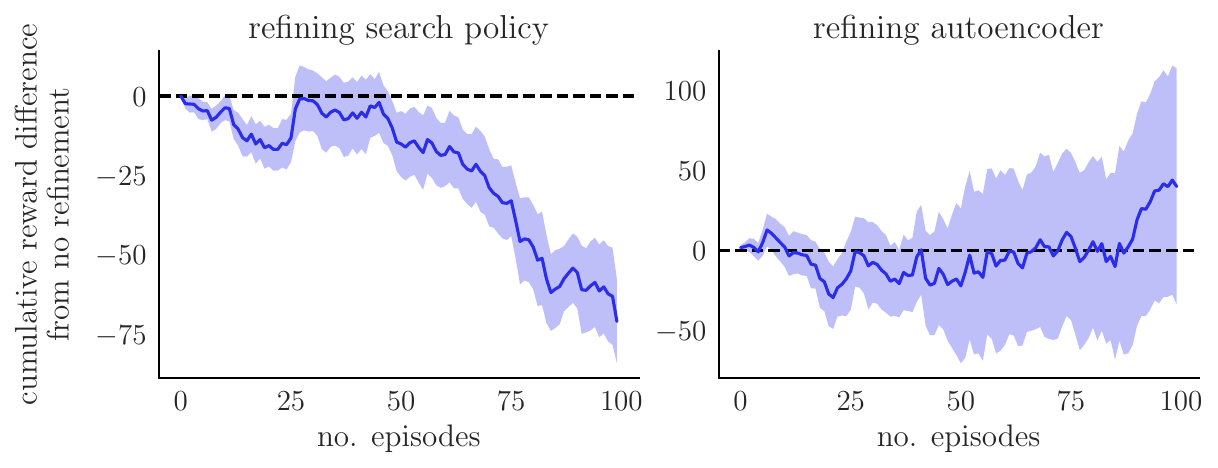}
\vskip -5pt
\caption{
Average cumulative difference in the number of targets found and standard errors between (\textbf{left}) updating the policy network using \acro{REINFORCE} or (\textbf{right}) updating the autoencoder producing the representation of the candidates vs.\ performing no updates.
}
\label{fig:reward_diff_refine}
\end{figure}

\cref{fig:reward_diff_refine} shows the value of each of the two update schemes as the cumulative difference in the number of targets found between each scheme compared to performing no updates (both the search policy and the representation of the data points are kept fixed) throughout 100 search episodes across 10 repeats.
Surprisingly, attempting to further refine the search policy using \acro{REINFORCE} actually hurts performance, resulting in an increasing gap in reward between the initial policy and the one continually updated.
On the other hand, we see that updating the initial unsupervised \acro{VAE} to account for the observed labels yields an improvement in performance on average, but this improvement is not consistent across the 10 repeats.
Overall, we show the difficulty in further updating our trained search policy using real experiences under repeated searches, and hypothesize that more sophisticated reinforcement learning procedures such as the double Q-learning algorithm \cite{mnih2015human} are needed to improve learning, which we leave as future work.

\section{Limitations}
\label{sec:limit}

\cref{sec:experiments} demonstrates that our policy \acro{ANS}, trained with imitation learning, cannot perfectly capture the search strategy of the state-of-the-art \acro{ENS} by \citet{jiang2017efficient} and is outperformed by the policy.
We view improving the architecture of our policy network as well as the state representation to enable more effective learning as a promising future direction.
For example, as examined in \citet{liu2020task}, a transformer-based network with beneficial input permutation invariance properties can learn from data sets of different sizes, which could also prove useful in \acro{AS}.
This network architecture can be combined with a better training strategy, as mentioned in \cref{sec:more_experiments}, to potentially yield comparable performance as \acro{ENS} or even to outperform it.

\section{Broader impact}
\label{sec:impact}

The development of an efficient \acro{AS} algorithm through imitation learning has significant potential to positively impact various fields where computational efficiency is paramount.
By reducing the superlinear computational complexity of the state-of-the-art policy \acro{ENS}, our approach enables the application of \acro{AS} in significantly larger data sets. 
This scalability is essential for industries and research fields that deal with vast amounts of data, including genomics, astronomy, and environmental monitoring, among others.
The ability to achieve competitive performance at a fraction of the cost also has substantial economic implications.
Organizations can deploy high-performance \acro{AS} solutions without the need for extensive computational resources, making advanced data analysis more accessible and affordable.
Since \acro{AS} aims to identify as many targets as possible, the collected data set may end up unbalanced towards the positives.
It is important for the user to ensure that maximizing the number of labeled targets accurately reflects their objective, and that the collected data are used by downstream tasks that are not negatively affected by this imbalance.

\end{document}